\documentclass{article}

\usepackage{microtype}
\usepackage{graphicx}
\usepackage{subcaption}
\usepackage{booktabs}
\usepackage{url}
\usepackage{float}

\usepackage{hyperref}

\usepackage[accepted]{icml2026}

\usepackage[utf8]{inputenc}
\usepackage[T1]{fontenc}
\usepackage{amsmath}
\usepackage{amssymb}
\usepackage{amsfonts}
\usepackage{mathtools}
\usepackage{amsthm}
\usepackage{nicefrac}
\usepackage{xcolor}
\usepackage[most]{tcolorbox}
\usepackage[capitalize,noabbrev]{cleveref}

\tcbset{
  finding/.style={
    colback=gray!8,
    colframe=black,
    fonttitle=\bfseries,
    title=Takeaway~#1,
    sharp corners,
    boxrule=0.4pt,
    left=6pt, right=6pt, top=4pt, bottom=4pt
  }
}

\theoremstyle{plain}

\theoremstyle{definition}

\theoremstyle{remark}

\newcommand{\method}{\textsc{AdaptGNS}}
\newcommand{\sand}{\textsc{Sand}}
\newcommand{\water}{\textsc{WaterDrop}}

\icmltitlerunning{Adaptive Interaction Graphs for Particle Simulation}

\begin{document}

\twocolumn[
  \icmltitle{Adaptive Interaction Graphs for Particle Simulation}

  \begin{icmlauthorlist}
    \icmlauthor{Aiden Zhou}{yale}
  \end{icmlauthorlist}

  \icmlaffiliation{yale}{Yale University, New Haven, Connecticut, USA}
  \icmlcorrespondingauthor{Aiden Zhou}{aiden.zhou@yale.edu}

  \icmlkeywords{Machine Learning, Graph Neural Networks, Particle Simulation, Uncertainty, Adaptive Graphs}

  \vskip 0.3in
]

\printAffiliationsAndNotice{}

\begin{abstract}
Learned particle simulators based on graph neural networks achieve strong one-step accuracy, but errors compound over long horizons. An underexplored variable is the interaction graph: existing methods fix its topology via $k$-nearest neighbors or a static radius rule, regardless of local model confidence. We propose making this graph adaptive: a per-particle variance head, trained jointly with the acceleration head under a
heteroscedastic Gaussian NLL loss, drives a trajectory in which high-uncertainty particles receive an expanded neighborhood. This is done at little extra inference cost by using the previous step's uncertainty estimate. A key discovery is that the variance head learns a meaningful notion of uncertainty: high-variance particles concentrate near complex regions, such as splash zones or free surfaces. When this signal drives graph topology, the resulting \textsc{AdaptGNS} simulator achieves a strict Pareto improvement on \textsc{WaterDrop} and a modest gain on \textsc{Sand}. Given the model's stronger performance on \textsc{WaterDrop}, we hypothesize that adaptive graphs are most useful when complexity is concentrated in space. Our code can be found at \href{https://github.com/aidenzhou8/AdaptGNS}{https://github.com/aidenzhou8/AdaptGNS}.

\end{abstract}

\section{Introduction} 

\paragraph{Problem statement.}
Learned simulators based on graph neural networks (GNNs) have shown strong one-step accuracy on particle dynamics tasks, but are hurt by compounding error over many steps \cite{gns2020}. A key architectural choice in such systems is the \textbf{interaction graph}, which decides what particles exchange messages at each time step. Existing methods fix this topology using static geometric heuristics, such as a radius rule \cite{gns2020} or a learned edge mask \cite{kipf2018}, regardless of the model's local confidence.

We propose \textsc{AdaptGNS}, an \textbf{uncertainty-adaptive interaction graph} that selectively expands edges per timestep based on the simulator's predicted uncertainty. By providing greater relational capacity where the model is least confident — and leaving the default sparse connectivity in place everywhere else — \textsc{AdaptGNS} aims to make graph network simulators (GNSs) more stable, but also to preserve the speed and efficiency that motivate using a learned simulator in the first place.

\paragraph{Motivation and significance.}
Long-horizon physical simulation is key to a wide range of subjects, such as robotics, molecular dynamics, and world models. Learned simulators can provide much faster inference than classical solvers, 
with reported speedups of up to 5{,}000$\times$ over the Material Point Method (MPM) on granular flow  simulations \cite{gns_joss}. More broadly, neural surrogate models for fluid simulation have been reported
to reduce simulation cost by orders of magnitude relative to conventional
numerical solvers \citep{lino2023deepfluid}. However, this benefit is lost if errors accumulate over many steps. Enhancing consistency without increasing average compute per step would make learned simulators a more viable drop-in replacement.

A second motivation comes from the physical observation that \textbf{uncertainty is rarely uniform}. When two fluids crash against each other, the unpredictable dynamics happen inside a ``splash zone,'' whereas the rest of the pool sits quietly. Our adaptive graph methodology encodes this intuition.

\paragraph{Learning objective.}
Our task is per-particle acceleration prediction at each timestep, with the predicted values integrated forward to produce a complete simulation trajectory. Errors at each step compound over many steps, and our goal is to reduce this error growth relative to baseline models. A secondary objective trains a \textbf{per-particle uncertainty head} — a scalar variance estimate $\sigma_i$ — that drives the adaptive graph policy at each step. We report \textbf{MSE at step 200 (MSE@200)} as a standard measure of accuracy and \textbf{average edges per step} as a hardware-agnostic proxy for efficiency. Note that each processor layer of a GNS does $O(|E|)$ work, so compute scales with edge count. 

\paragraph{Summary of results.} \textsc{AdaptGNS} achieves a strict Pareto improvement on \textsc{WaterDrop} (20\% fewer edges and lower MSE@200 than fixed $k=5$), but only a modest gain on \textsc{Sand} (12\% lower MSE@200 than fixed $k=5$, using 10\% fewer edges than $k=10$).\footnote{$k \approx$ the average number of neighbors per particle at the given connectivity radius $r$. We use this notation throughout, but be aware that $r$ is fixed; neighbor count varies with local density.} The story becomes clearer once we ablate for the choice of loss function: compared to a fixed GNS \textbf{trained on NLL loss}, the adaptive policy once more achieves a strong gain (21\% fewer edges and 8\% lower MSE@200). We also report the optimized hyperparameters found by grid sweep on \textsc{Sand} and \textsc{WaterDrop} ($\sigma$-percentile $= 70$, radius factor $= 1.267$) and several negative results.

\section{Dataset}

\paragraph{Description.}
We use two distinct particle simulation datasets. \textsc{Sand} is taken from \cite{gns_joss}, hosted on DesignSafe-CI (DOI: \texttt{10.17603/ds2-0phb-dg64}), and generated with the Taichi-MPM solver. \textsc{WaterDrop} is taken from \cite{gns2020} and generated with a Smoothed Particle Hydrodynamics (SPH) solver. Note that \textsc{WaterDrop} trajectories are given as TFRecords, so we provide a script to convert the dataset to a standard \texttt{.npz} format. These two materials were chosen to evaluate adaptive graphs across both fluid and granular flow regimes.

\begin{figure}[tb]
    \centering
    \includegraphics[width=\linewidth]{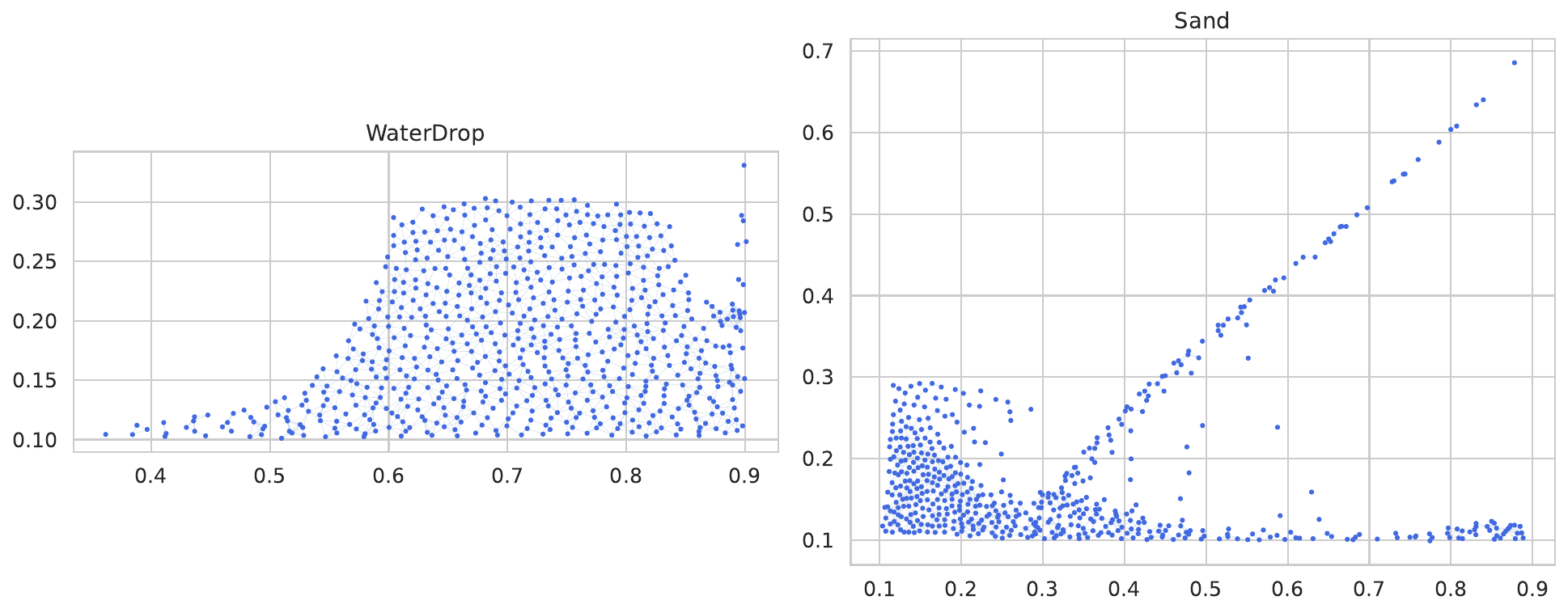}
    \caption{Interaction graph snapshots at Step 100 for \textsc{WaterDrop} (left) and \textsc{Sand} (right).}
    \label{fig:graph_snapshot}
\end{figure}

\begin{table}[tb]
  \caption{Data splits and rollout lengths.}
  \label{tab:data_splits}
  \centering
  \begin{tabular}{lrrrr}
    \toprule
    Dataset & Train & Val. & Test & Steps \\
    \midrule
    \sand{} & 1{,}000 & 100 & 30 & 320 \\
    \water{} & 1{,}000 & 30 & 30 & 1{,}000 \\
    \bottomrule
  \end{tabular}
\end{table}

\paragraph{Graph statistics.} The edges of a standard interaction graph are constructed at each step between particles separated by at most $r = 0.015$. Table~\ref{tab:graph_stats} reports graph statistics taken at Step 100, averaged across trajectories. Both datasets produce sparse graphs with moderate average degree.

\begin{table*}[tb]
  \caption{Graph statistics at Step 100, averaged over 30 test trajectories.}
  \label{tab:graph_stats}
  \centering
  \begin{tabular}{lcccccc}
    \toprule
    Dataset & Trajs & Nodes & Edges & Density & Avg Degree & Avg Clustering \\
    \midrule
    \textsc{WaterDrop} & 30  & $567 \pm 200$  & $1550 \pm 583$  
              & $1.08 \times 10^{-2}$ & $5.42 \pm 0.22$ & $0.494 \pm 0.036$ \\
    \textsc{Sand}      & 30 & $1270 \pm 503$ & $4324 \pm 1864$ 
              & $7.35 \times 10^{-3}$ & $6.76 \pm 1.01$ & $0.500 \pm 0.034$ \\
    \bottomrule
  \end{tabular}
\end{table*}

The temporal evolution of graph 
statistics reveals varying dynamics: in \textsc{WaterDrop}, 
edge count and average degree rise at impact and taper as particles disperse, whereas in \textsc{Sand} the graph undergoes a sharp slump around Step 50 before recovering to a stable plateau as particles settle. These signatures are central to understanding the performance gap between datasets, discussed in Section 5.2.

\section{Related Work}

\paragraph{Graph network simulators.} The idea of representing physical systems as graphs was introduced by \cite{battaglia2016}, who developed the relational inductive bias that subsequent work extends upon. \cite{gns2020} scaled this paradigm to large particle 
systems via radius-based interaction graphs and noise-augmented training. This work forms the 
primary baseline and codebase for our project. A central restriction shared by both 
works is that the interaction graph is fixed by geometric rules, which do not regard local dynamics. The phenomenon of ``over-squashing'' provides a theoretical motivation: fixed sparse graphs can restrict the information exchanged across bottleneck regions on a per-step basis \cite{topping2022}. In autoregressive simulators, such representational constraints may interact with compounding error, motivating adaptive distribution of edges in complex and uncertain regions.

\paragraph{Adaptive predecessors.} The most straightforward try to learn the graph structure itself came from \cite{kipf2018}, whose Neural Relational Inference (NRI) derives a discrete interaction graph via a VAE encoder over particle histories. However, NRI produces a graph that is fixed at inference, and incurs significant overhead. More recently, AdaMeshNet \cite{adameshnet2025} proposes 
per-step rewiring for mesh-based GNNs using geometric heuristics, outperforming 
static baselines. Our work diverges from both: 
we target point-cloud particle systems rather than meshes, use no separate encoder, and use the model's own predicted uncertainty $\sigma_i$ as 
the adaptation signal.

\paragraph{Physics priors and non-graph alternatives.}
Several recent approaches look to enhance long-horizon simulation without adapting a particle interaction graph. Neural SPH augments learned Lagrangian fluid simulators with components taken from Smoothed Particle Hydrodynamics (SPH) \cite{neural_sph2024}. NeuralMPM uses a hybrid particle-grid representation: particles are interpolated to a regular grid, updated with image-to-image networks, and interpolated back \cite{neuralmpm2024}. Lastly, Transformer-based particle simulators such as FluidFormer combine local convolutions with global attention to capture long-range dependencies \cite{fluidformer2025}. These methods address the same problem of long-horizon consistency, but through physics priors, grid-based emulation, or global attention. \method{} is orthogonal to these directions, and could in principle be integrated with stronger message functions or physics-informed losses. 
\section{Methodology}

\paragraph{Variance head.}
We extend a standard GNS with a \textbf{variance prediction head}: alongside the acceleration output, the network predicts a scalar uncertainty $\sigma_i$ per particle. This is done via a two-layer MLP on the final node embeddings, using a softplus activation to enforce positivity: \[
\sigma_i^2 = \operatorname{softplus}(f_{\theta}(h_i)).
\]
This change adds fewer than 20,000 parameters to the standard GNS, so the architectural overhead is insignificant relative to the base simulator.

\paragraph{Heteroscedastic NLL loss.} Both acceleration and variance heads are trained jointly under a Gaussian NLL loss:
\begin{equation}
  \mathcal{L}
  =
  \frac{1}{N}\sum_{i=1}^{N}
  \left[
  \frac{\|\hat{\mathbf a}_i-\mathbf a_i\|_2^2}{2\sigma_i^2}
  +
  \frac{d}{2}\log \sigma_i^2
  \right],
  \label{eq:nll}
\end{equation}

where $d=2$ is the acceleration dimension. NLL ensures the model is rewarded for concentrating uncertainty where error is large and punished for inflating it elsewhere. 

\paragraph{Adaptive graph construction loop.} Our core contribution is an inference-time loop that uses predicted uncertainty to modify the graph topology. A naive implementation would run a GNN forward pass, extract
$\sigma_i$, rebuild the graph, and then run a second forward pass to compute
the final acceleration. However, this doubles inference cost. Instead, we use a \textbf{single-pass lagged-$\sigma$} design. At step
$t$, the graph is constructed using the previous step's uncertainty estimates
$\sigma_i^{(t-1)}$. The forward pass at step $t$ then produces both the next
position and a fresh uncertainty estimate $\sigma_i^{(t)}$ for
step $t + 1$. Concretely, the adaptive rule is:
\begin{enumerate}
    \item Construct the standard interaction graph.
    \item Compute the per-step $\tau$-th percentile of $\{\sigma_i\}_{i=1}^N$ and mark particles above this percentile as high uncertainty.
    \item Add extra edges from a larger radius
    $r_{\text{large}} = \texttt{rad\_factor} \cdot r$ for edges incident to one or more high-uncertainty particles.
\end{enumerate}

\paragraph{Hyperparameters.} Using a grid sweep over validation set trajectories, we selected values of $r=0.015$, $\texttt{rad\_factor}=1.267$, and $\tau=70$. The expanded radius roughly matches $r=0.019$, so high-$\sigma$ particles receive the same connectivity as the strongest fixed-graph baseline.

\section{Experiments and Results}

\paragraph{Baselines and Ablations.} We compare \textsc{AdaptGNS} to five models on both \textsc{WaterDrop} and \textsc{Sand}. The naive comparison is to a \textbf{per-particle MLP}, which measures the value of relational inductive bias. Far more reasonable models are a \textbf{fixed-graph GNS with $k = 5$} (or $r = 0.015$), which serves as the primary comparison, and a \textbf{fixed-graph GNS with $k = 10$} (or $r = 0.019$), which gives an upper bound on the benefit of denser interaction graphs. 

We also conduct two ablation studies. \textbf{GNS ($k$-mid)} uses an intermediate radius ($r=0.0176$) matched to \textsc{AdaptGNS}'s mean edge count, isolating whether our gains come from adaptive topology or just from having more edges on average.\footnote{Note that this ablation was not conducted for \textsc{WaterDrop}. On \textsc{Sand}, $k_\text{mid}$ was constructed by interpolating in $r^2$ between the $k = 5$ and $k = 10$ edge counts to match \textsc{AdaptGNS}'s budget, giving $r \approx 0.0176$. On \textsc{WaterDrop}, \textsc{AdaptGNS} already uses \textbf{fewer} edges than fixed $k = 5$, so no equivalent comparison exists.}
\textbf{\textsc{AdaptGNS} fixed-graph uses NLL} but with fixed $r=0.015$, isolating the contribution of the adaptive loop from the changed loss function.

\subsection{Quantitative Results}

Our first question is whether the variance head reveals physical structure. If $\sigma_i$ is just noise, a graph policy that depends on it is doubtful to succeed. We find evidence that $\sigma_i$ acts as an accurate rank-based signal, even though it is not perfectly calibrated in absolute value.

\begin{tcolorbox}[finding=1]
\textbf{As shown in Figure 2, the network correctly infers what ``uncertainty'' means in the context of physical simulation, purely as a byproduct of the learning objective.} This is supported by our calculations in Table 3. On \textsc{Sand}, Bins 0–7 are tightly calibrated, with $|\Delta| \leq 0.06$. The total ECE is large only because the model severely underestimates its error on the hardest particles (Bins 8 and 9). Results are weaker on \textsc{WaterDrop}, but conform to the same trend. As our adaptive methodology thresholds at a percentile rank, this does not pose a huge obstacle.
\end{tcolorbox}

\begin{figure}[tb]
    \centering
    \includegraphics[width=\linewidth]{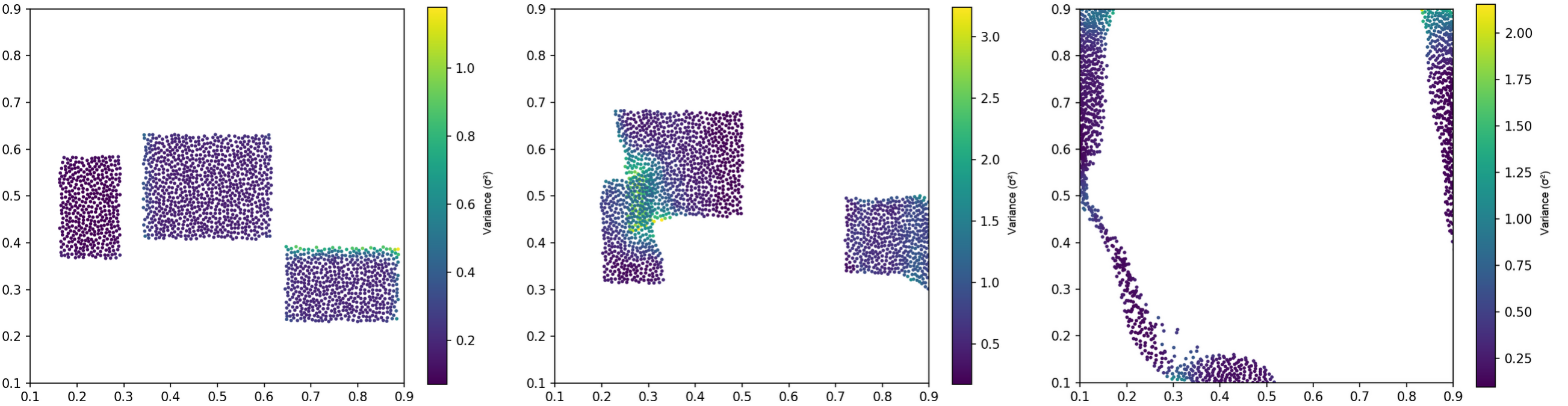}
    \caption{Per-particle uncertainty $\sigma_i$ at steps 1, 10, and 50 of a \textsc{Sand} trajectory. High-$\sigma$ particles are concentrated in impact zones, free surfaces, and regions of rapid velocity change.}
    \label{fig:sigma_spatial}
\end{figure}

\begin{table*}[tb]
  \caption{Per-decile calibration on \textsc{Sand} and \textsc{WaterDrop}.}
  \label{tab:calibration_sand_wd}
  \centering
  \footnotesize
  \begin{tabular}{c|ccc|ccc}
    \toprule
    & \multicolumn{3}{c|}{\textsc{Sand}} & \multicolumn{3}{c}{\textsc{WaterDrop}} \\
    Bin & mean SE & $2\hat{\sigma}^2$ & $|\Delta|$ & mean SE & $2\hat{\sigma}^2$ & $|\Delta|$ \\
    \midrule
    0 & 0.023 & 0.009 & 0.014 & 0.543 & 0.003 & 0.540 \\
    1 & 0.025 & 0.014 & 0.011 & 0.540 & 0.004 & 0.537 \\
    2 & 0.024 & 0.018 & 0.006 & 0.131 & 0.009 & 0.121 \\
    3 & 0.026 & 0.023 & 0.003 & 0.081 & 0.015 & 0.066 \\
    4 & 0.033 & 0.029 & 0.004 & 0.208 & 0.022 & 0.186 \\
    5 & 0.046 & 0.038 & 0.008 & 0.495 & 0.032 & 0.463 \\
    6 & 0.067 & 0.051 & 0.016 & 1.386 & 0.048 & 1.339 \\
    7 & 0.141 & 0.084 & 0.057 & 3.167 & 0.065 & 3.101 \\
    8 & 0.560 & 0.195 & 0.365 & 5.739 & 0.094 & 5.645 \\
    9 & 15.482 & 1.882 & 13.599 & 17.878 & 0.288 & 17.590 \\
    \midrule
    \multicolumn{1}{l}{Total ECE} & & & 1.408 & & & 2.959 \\
    \bottomrule
  \end{tabular}
\end{table*}

Next, we evaluate \textsc{AdaptGNS} on \textsc{Sand} and \textsc{WaterDrop}. See Table 4 for a summary of results and Figure 3 for more nuanced dynamics.

\paragraph{\textsc{Sand}.} Comparing \textsc{AdaptGNS} to its fixed-graph counterpart (using NLL and $r = 0.015$), the adaptive policy \textbf{reduces edges by 21\% and
lowers MSE@200 by 8\% (0.0417 vs. 0.0382)}.
This is the cleanest comparison: it isolates the value of placing edges
purposely rather than uniformly.
Second, comparing \textsc{AdaptGNS} to $k_\text{mid}$ ($r = 0.0176$) at matched edge counts
(${\sim}23$,000 edges each), \textbf{the model using MSE wins (0.0355 vs.\ 0.0382)}. The benefit of MSE as an easier learning objective is made clear by the NLL model at $r = 0.0176$. Although it also uses a larger fixed radius, its MSE@200 is much worse (0.0422).

\paragraph{\textsc{WaterDrop}.}
The result here is a strong advance: \textsc{AdaptGNS} \textbf{uses 20\%
fewer edges than $k = 5$ (7,821 vs.\ 9,741) and also achieves lower MSE@200
(0.0412 vs.\ 0.0414).}
Even compared to $k = 10$, the tradeoff is 36\% fewer edges for a 6\% relative MSE@200
increase. In contrast to \textsc{Sand}, the decrease in performance caused by using NLL instead of MSE is close to zero.

\begin{table*}[tb]
  \caption{Results on \textsc{Sand} (320 steps) and \textsc{WaterDrop} (1000 steps).}
  \label{tab:rollouts_sand_wd_mse200}
  \centering
  \footnotesize
  \setlength{\tabcolsep}{5pt}
  \begin{minipage}[t]{0.48\textwidth}
    \centering
    \textbf{\textsc{Sand}}\\[4pt]
    \begin{tabular}{@{}lrr@{}}
      \toprule
      Model & Edges/step & MSE@200 \\
      \midrule
      MLP & N/A & 0.121 \\
      $k =5$ ($r = 0.015$) & 18{,}682 & 0.0434 \\
      $k_{\mathrm{mid}}$ ($r = 0.0176$) & 23{,}498 & 0.0355 \\
      $k=10$ ($r = 0.019$) & 25{,}516 & \textbf{0.0285} \\
      NLL ($r = 0.015$) & 28{,}968 & 0.0417 \\
      NLL ($r = 0.0176$) & 28{,}768 & 0.0422 \\
      \textbf{AdaptGNS} & \textbf{22{,}958} & 0.0382 \\
      \bottomrule
    \end{tabular}
  \end{minipage}\hfill
  \begin{minipage}[t]{0.48\textwidth}
    \centering
    \textbf{\textsc{WaterDrop}}\\[4pt]
    \begin{tabular}{@{}lrr@{}}
      \toprule
      Model & Edges/step & MSE@200 \\
      \midrule
      MLP & N/A & 0.207 \\
      $k=5$ ($r = 0.015$) & 9{,}741 & 0.0414 \\
      $k=10$ ($r = 0.019$) & 12{,}212 & \textbf{0.0388} \\
      NLL ($r = 0.015$) & 7{,}875 & 0.0417 \\
      \textbf{AdaptGNS} & \textbf{7{,}821} & 0.0412 \\
      \bottomrule
    \end{tabular}
  \end{minipage}
\end{table*}

\begin{figure*}[tb]
    \centering
    \includegraphics[width=\linewidth]{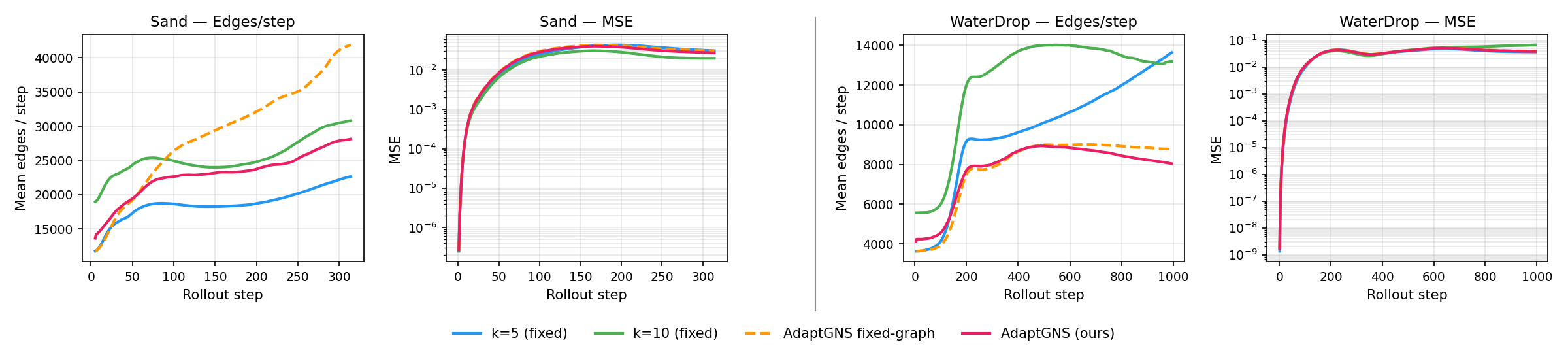}
    \caption{Per-step dynamics (edge count and MSE) on \textsc{Sand} and \textsc{WaterDrop}.}
    \label{fig:edges_mse_dynamics}
\end{figure*}

\begin{tcolorbox}[finding=2]
\textbf{\textsc{AdaptGNS} strictly beats the same NLL model without the adaptive graph mechanism on both datasets.} On \textsc{WaterDrop}, it also achieves a strong advance over every fixed-graph comparison. Due to its harder learning objective, \textsc{AdaptGNS} does not yet defeat the best MSE-trained models on \textsc{Sand}.
\end{tcolorbox}

\subsection{Analysis and Discussion}

\paragraph{Why \textsc{WaterDrop} and not \textsc{Sand}?}
The performance gap between datasets can be attributed to two physical properties. First, \textbf{graph density evolves differently}. Most granular trajectories end up packing particles into an increasingly dense heap. A droplet impact does the opposite, dispersing particles and keeping the graph sparse for the rest of the trajectory. Second, \textbf{uncertainty is more heterogeneous in \textsc{WaterDrop}}: a splash creates local complexity but is mostly surrounded by quiet, giving the variance head plenty of contrastive signal. In \textsc{Sand}, nearly every particle undergoes the same dynamic, as was previously observed in our exploratory data analysis.

    \paragraph{MSE vs. NLL.} A second question concerns the model's objective. The variance head requires a learning objective that integrates uncertainty, but NLL can be harder for raw acceleration accuracy than MSE. This is visible on \sand{}, where the best MSE baselines outperform NLL models. Hence, a future goal is to decouple acceleration / force prediction from uncertainty ranking; for instance, by using MSE loss along with a subsidiary ranking or calibration loss for the variance head.

\paragraph{Edge count is a noisy proxy for compute.}
The radius graph at step $t$ is constructed from the model's predicted positions. Hence, a key observation is that \textbf{worse models can drift into denser configurations and end up using more edges}. This is shown by our results in Table 4 and Figure 3, particularly for the two NLL \textsc{Sand} models.

\section{Limitations and Future Work}

Several restrictions should be considered when interpreting this study, and addressed in future work.

\paragraph{Runtime measurement.}
We report average edge count because it tracks the principal cost of a GNN forward pass, but it is not a complete runtime metric. Adaptive graph construction searches over an expanded radius for selected particles. The overhead this process creates should be quantified in future work.

\paragraph{Lagged uncertainty.}
The single-pass design uses $\sigma^{(t-1)}$ to construct the graph at step $t$. This avoids a second forward pass but may react too late during abrupt transformations. A next-step predictor or a physical prior based on acceleration or velocity divergence could reduce this lag.

\paragraph{Scope.}
The present experiments cover one SPH fluid dataset and one MPM granular dataset. Broader evaluation on more materials, 3D contexts, and hybrid rigid--fluid scenes is needed before drawing general conclusions.

\paragraph{Broader domains.}
Uncertainty-guided graph adaptation may also prove useful beyond particle simulation. Natural extensions include molecular dynamics, where reactive sites or conformational changes may benefit from larger local neighborhoods; mesh-based PDE solvers, where shocks or boundary layers motivate adaptive refinement; and robotics, where local interactions often dominate trajectory error.
\section{Conclusion}

In this paper, we explored making the interaction graph of a learned particle simulator adaptive — sparse where the model is confident, denser where it is not — without paying the cost of two GNN computations per step. We contribute a single-pass adaptive mechanism driven by a per-particle variance head, which is trained jointly with the acceleration head under NLL loss. We observe that $\sigma_i$ corresponds to real physical uncertainty, and that \textsc{AdaptGNS}, driven by this signal, achieves modest but notable results on the \textsc{WaterDrop} and \textsc{Sand} datasets.

Future work should target variance-aware learning objectives that preserve MSE accuracy, a next-step uncertainty predictor (solving both lag and calibration concerns) and a learnable per-particle expansion radius. We also intend to test \textsc{AdaptGNS} on a broader range of physical regimes and on longer-horizon tasks, and to conduct rigorous multi-seed evaluations. These are needed to more carefully examine the general usefulness of adaptive interaction graphs for learned particle simulation. 

\section*{Impact Statement}
This paper presents work whose goal is to refine learned physical simulation. Potential societal consequences are tied to downstream uses of faster and more accurate simulation systems. We do not identify additional societal risks beyond those typically associated with machine learning research.

\bibliography{references}
\bibliographystyle{icml2026}

\end{document}